\documentclass[conference,10pt]{IEEEtran}
\usepackage{subfigure}
\usepackage{color}
\usepackage{xcolor}
\usepackage{comment}
\usepackage{algorithm,algpseudocode}
\usepackage{rotating}
\usepackage{multirow}
\usepackage{enumitem}
\usepackage{url}
\usepackage{xspace}
\usepackage{subfigure}
\usepackage{amsmath}
\usepackage{tabularx}  
\usepackage{graphicx}
\usepackage{amsfonts}

\newcolumntype{C}[1]{>{\centering\arraybackslash}p{#1}}

\newcommand{\eg}{\textit{e.g.},\xspace}
\newcommand{\ie}{\textit{i.e.},\xspace}
\definecolor{lightgreen}{rgb}{0.56, 0.93, 0.56}
\begin{document}

\title{Multi-Service Mobile Traffic Forecasting via Convolutional Long Short-Term Memories}

\author{\IEEEauthorblockN{Chaoyun Zhang}
\IEEEauthorblockA{\textit{University of Edinburgh} \\
chaoyun.zhang@ed.ac.uk\vspace*{-8pt}}
\and
\IEEEauthorblockN{Marco Fiore}
\IEEEauthorblockA{\textit{CNR-IEIIT} \\
marco.fiore@ieiit.cnr.it\vspace*{-8pt}}
\and
\IEEEauthorblockN{Paul Patras}
\IEEEauthorblockA{\textit{University of Edinburgh} \\
paul.patras@ed.ac.uk\vspace*{-8pt}}
}

\maketitle
\begin{abstract}
Network slicing is increasingly used to partition network infrastructure between different mobile services. Precise service-wise mobile traffic forecasting becomes essential in this context, as mobile operators seek to pre-allocate resources to each slice in advance, to meet the distinct requirements of individual services. This paper attacks the problem of multi-service mobile traffic forecasting using a sequence-to-sequence (S2S) learning paradigm and convolutional long short-term memories (ConvLSTMs). The proposed architecture is designed so as to effectively extract complex spatiotemporal features of mobile network traffic and predict with high accuracy the future demands for individual services at city scale. We conduct experiments on a mobile traffic dataset collected in a large European metropolis, demonstrating that the proposed S2S-ConvLSTM can forecast the mobile traffic volume produced by tens of different services in advance of up to one hour, by just using measurements taken during the past hour. In particular, our solution achieves mean absolute errors (MAE) at antenna level that are below 13KBps, outperforming other deep learning approaches by up to~31.2\%. 
\end{abstract}

\begin{IEEEkeywords}
Mobile traffic forecasting; deep learning; convolutional long short-term memory.
\end{IEEEkeywords}

\section{Introduction}\label{sec:intro}
The rapid development of 5G networks and IoT technology is triggering a surge in mobile traffic consumption globally. As a matter of fact, the latest estimations indicate that 77.5 exabytes of mobile data will be consumed per month by 2022, which accounts for 71\% of the total IP traffic \cite{cisco2017}. At the same time, digital services continue to diversify and demand often conflicting performance guarantees (\eg low latency vs. high throughput vs. high reliability). Precision traffic engineering thus becomes increasingly important, as operator must be able to allocate resources intelligently for each service and anticipate individual future demands.
Network slicing is a first step towards addressing these challenges, enabling to logically isolate network infrastructure on a per-service basis. Allocating sufficient resources to a certain slice however requires precise measurements and detailed real-time analysis of mobile traffic~\cite{bega2019deepcog}. Achieving this is computationally expensive~\cite{naboulsi2016large} and relies heavily on specialized equipment (\eg measurement probes~\cite{keysight}). 

To alleviate these issues, mobile operators seek to forecast future mobile traffic consumption, so as to be able to react to changes in traffic volume in advance. Driven by recent progress in advanced parallel computing, deep learning is becoming increasingly important in this area~\cite{zhang2018deep}. In particular deep learning based predictors can minimize the effort devoted to feature engineering and automatically extract spatiotemporal dependencies inherent to mobile traffic \cite{zhang2017zipnet}. However, existing deep learning-based forecasting approaches largely deliver predictions only on traffic aggregates (\eg \cite{zhang2018long, wang2018spatio}). Hence, such tools have limited applicability to network slicing, where precise information of the future traffic demand of individual services is required.

To achieve precise mobile traffic forecasting in support of network slicing, in this paper we propose to leverage sequence-to-sequence learning \cite{sutskever2014sequence} and convolutional long short-term memories (ConvLSTMs) \cite{xingjian2015convolutional}, targeting network-wide traffic prediction over a broad range of mobile services. The ConvLSTM structure incorporates convolution operations into long short-term memory (LSTM), enabling it to capture dependencies in both spatial and temporal dimensions, which are routinely encountered in mobile traffic. By combining this with sequence-to-sequence (S2S) learning, our architecture performs end-to-end forecasting with minimal assumptions on the structure of traffic measurement time series. 

We conduct experiments with a large-scale mobile traffic dataset consisting of detailed measurements collected by a major operator at nearly 800 antennas distributed across a metropolitan area in Europe. Test results over 17 consecutive days demonstrate that our proposed deep learning architecture can perform 1-hour mobile traffic forecasting with high accuracy for 36 distinct services. Specifically, our solution achieves a mean absolute error (MAE) of 13KBps per service per antenna and up to 9.63\% prediction error in terms of normalized MAE (NMAE), outperforming deep learning baselines by up to 31.2\%. In addition, grouping services into 8 main categories leads to a NMAE that is as low as 3.79\%. These results confirm that our proposal can be used as a reliable tool for multi-service mobile traffic forecasting in support of resource management for network slicing.


\section{The Multi-service Forecasting Problem}
The objective of this work is to achieve precise forecasts of mobile data traffic consumption by individual services, at the level of multiple antennas distributed in a city, given sequences of historical traffic measurements. Formally, consider a geographical region where a set of antennas, $\mathcal{A}$, is deployed across different locations, to provide user coverage. Each antenna generates mobile data traffic continuously, which is consumed by a set of mobile services, $\mathcal{S}$. We denote the traffic demand accommodated by antenna $a\in\mathcal{A}$ for a specific service $s\in\mathcal{S}$ at time $t$ as $d_a^s(t)$. A network-level traffic snapshot for a service $s$ is denoted by $D^s(t)$, which gathers the demand of that service at all antennas distributed in the target region, \ie $D^s(t) = \left\{d_a^s(t) | a\in\mathcal{A}\right\}$. We further denote by $\mathcal{D}^\mathcal{S}(t)$ the set of snapshots at time $t$ for the full service set considered, \ie $\mathcal{D}^\mathcal{S}$ $ = \left\{D^s(t) | s\in\mathcal{S}\right\}$.

The mobile traffic forecasting problem of interest can be defined as inferring the traffic volumes in $K$ future mobile traffic snapshots given $T$ previous observations. From a machine learning perspective, this requires solving
\begin{equation}
\begin{aligned}
    &\widetilde{\mathcal{D}}^\mathcal{S}(t+1), \dots, \widetilde{\mathcal{D}}^\mathcal{S}(t+K) :=\\ &\mathop{\arg\max}\limits_{\mathcal{D}^\mathcal{S}(t), \dots, \mathcal{D}^\mathcal{S}(t+K)} p \big(\mathcal{D}^\mathcal{S}(t), \cdots, \mathcal{D}^\mathcal{S}(t+K)~|~\\
    &\qquad\qquad\qquad\qquad \mathcal{D}^\mathcal{S}(t-T+1),\dots,\mathcal{D}^\mathcal{S}(t)\big),
\label{eq:mtd}
\end{aligned}
\end{equation}
where $\widetilde{\mathcal{D}}^\mathcal{S}(t+k)$ denotes the traffic demands inferred 
over the service set $\mathcal{S}$ for all antennas in $\mathcal{A}$ at future time instances $t+k, 0 \leq k \leq K$. Solving this problem is not straightforward, as it requires to infer $K\times\mathcal{S}\times\mathcal{A}$ variables at a time, where intricate spatiotemporal correlations exist among these. To tackle this problem, we employ sequence-to-sequence learning and ConvLSTMs, which enables modelling complex traffic dependencies and achieving precise mobile traffic forecasting per individual services. We detail the proposed neural network architecture next.

\section{Sequence-to-Sequence ConvLSTM} 
\noindent\textbf{Convolutional long short-term memories (ConvLSTMs)} are a special class of recurrent neural networks that incorporate convolution operations into LSTM to extract correlations across both spatial and temporal dimensions. Compared to ``vanilla'' LSTM, the ConvLSTM replace the inner dense connections with convolution operations. This reduces significantly the number of parameters that must be configured in the model and enables to preserve important spatial correlations. This feature is particularly relevant to mobile data traffic modelling, as traffic snapshots exhibit embraces non-trivial spatio-temporal correlations~\cite{zhang2017zipnet}.

Given a sequence of 3-D inputs (\ie a sequence of mobile traffic snapshots) denoted $\mathbf{D} = \{ D(1), D(2),\dots, D(T) \}$, the operations performed by a single ConvLSTM are given in~(\ref{eq:convlstm})%
\footnote{To reduce clutter in the notation, we drop the superscripts that indicate a particular service in the remainder of the discussion.}.
Here `$\odot$' denotes the Hadamard product, `$*$' the 2-D convolution operator, and $\sigma(\cdot)$ is a sigmoid function.
\begin{equation}
\begin{aligned}
&i_t = \sigma(W_{xi}*D(t) + W_{hi}*H_{t-1}+W_{ci}\odot C_{t-1} + b_i),\\
&f_t = \sigma(W_{xf}*D(t) + W_{hf}*H_{t-1}+W_{cf}\odot C_{t-1}+b_f),\\
&C_t = f_t\odot C_{t-1} + i_t \odot \tanh(W_{xc}*D(t) + W_{hc}*H_{t-1}+b_c),\\
&o_t = \sigma(W_{xo}*D(t) + W_{ho}*H_{t-1}+W_{co} \odot C_t +b_o),\\
&H_t = o_t\odot \tanh(C_t).
\end{aligned}
\label{eq:convlstm}
\end{equation}
In the above, $W_{(\cdot \cdot)}$ and $b_{(\cdot)}$ denote the weights and biases of the model, which are learned during training. The inputs $D(t)$, cell outputs $C_t$, hidden states $H_t$, input gates $i_t$, forget gates $f_t$, and output gates $o_t$ in the \mbox{ConvLSTM's} inner structure are all 3-D tensors. The first two dimensions of the tensors form the spatial dimension, while the third is the number of feature maps, \ie traffic snapshots for different services $D^s(t)$. The ConvLSTM is configured with a set of gates, allowing the model to ``learn to forget" in both spatial and temporal dimensions. This is beneficial to capturing long-term dependencies in mobile traffic.\\

\begin{figure}[t!]
\centering\includegraphics[width=\columnwidth]{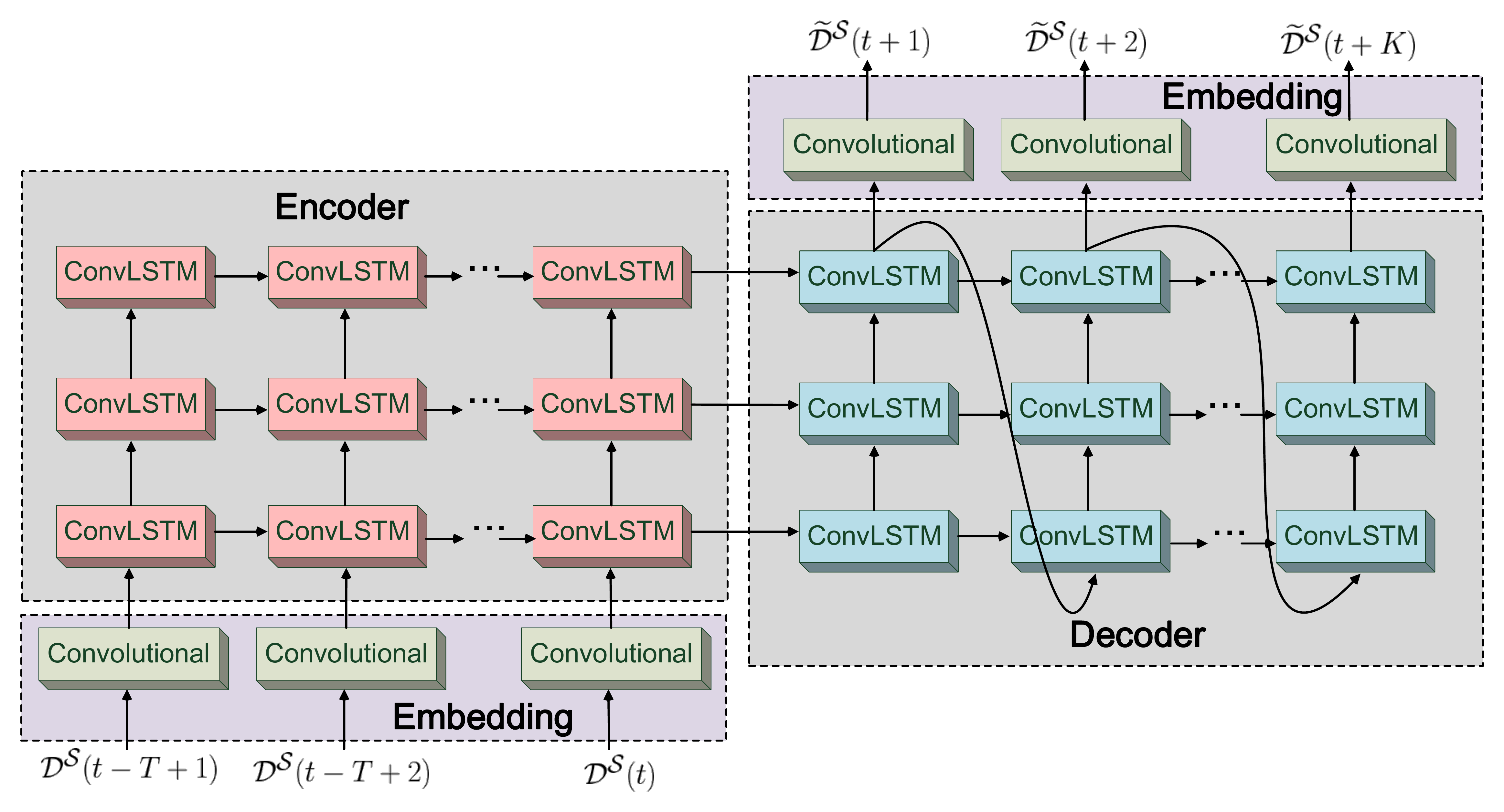}
\caption{The overall architecture of the ConvLSTM with sequence-to-sequence learning employed in this study for multi-service mobile traffic forecasting. \label{fig:convlstm}} 
\end{figure}

\noindent \textbf{Sequence-to-sequence learning} is a technique widely-applied to recurrent neural network (RNN) for machine translation tasks \cite{sutskever2014sequence}. This structure employs a RNN model to encode the input into a low-dimensional tensor. The decoder is another RNN model that decodes the tensor encoded into a sequence. The sequence-to-sequence architecture provides an end-to-end mapping from different sequences, which is particular well-suited for the mobile traffic forecasting problem, since essentially we seek to infer a sequence (\ie future mobile traffic demand) from another sequence (\ie mobile traffic measurements observation).\\

\noindent\textbf{S2S-ConvLSTM -- the overall architecture } of the proposed neural network for multi-service traffic forecasting is illustrated in Fig.~\ref{fig:convlstm}. Similar to the RGB channels of images, traffic snapshots of different services at a given time are treated as different convolutional channels. These snapshots will be first processed by convolutional embedding layers to meet the dimension requirement. Important spatio-temporal features of mobile traffic are distilled hierarchically through stacks of ConvLSTM layers, and are encoded and squeezed as the final states of the encoder at time $t$. Such states are delivered to a decoder, which is responsible for decoding the information encoded into predictions of future traffic volumes, through other convolutional embedding layers. The encoder-decoder architecture can be trained in an end-to-end manner.

\section{Mobile Traffic Dataset and Preprocessing}
\label{sub:data}



\begin{figure*}[t!]
\centering\includegraphics[width=2.1\columnwidth]{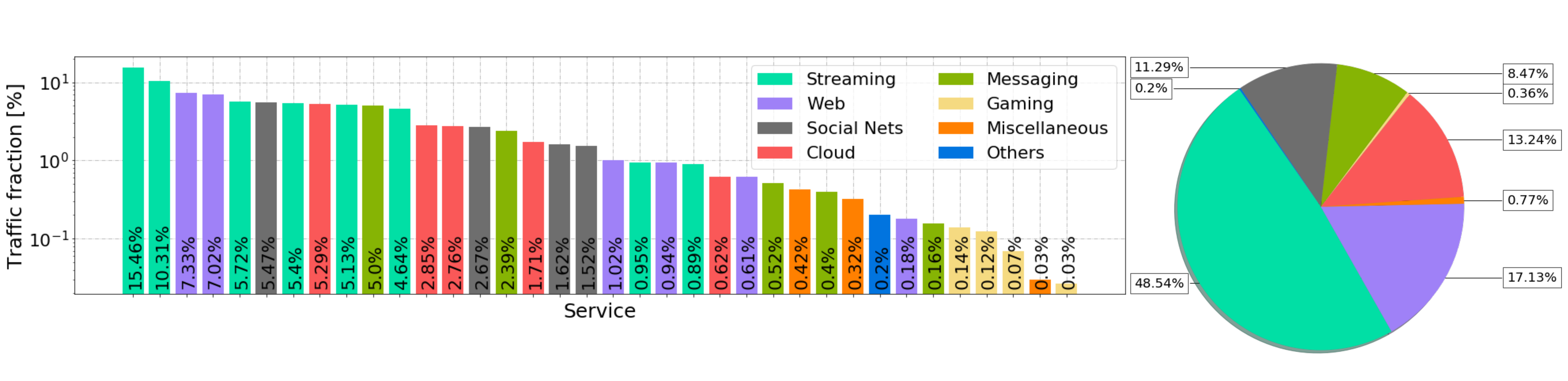}
\caption{Demands for services in the considered set $\mathcal{S}$. Fraction of the total traffic consumed by each mobile service (left) and each service category (right). \label{fig:traffic-stats}} 
\end{figure*}

To assess the performance of the proposed forecasting solution, we conduct experiments using a large-scale mobile traffic dataset collected by a major operator in a large European metropolitan area during 85 consecutive days. This produces 24,482 traffic snapshots for individual service. Each mobile traffic snapshot comprises the traffic demand accommodated by 792 antennas (\ie $|\mathcal{A}| = 792$) aggregated every 5 minutes. We filter antennas where active traffic flows exist at least 90\% of the time, in order to eliminate 2G antennas and those that have been decommissioned or are enabled only sporadically. Before feeding the traffic volume information to the model, uplink and downlink traffic is combined at each antenna. The dataset contains traffic consumption information for individual mobile services.

All data is collected at the network Packet Gateway (PGW) level and classified using deep packet inspection. This allows to identify traffic flow usage of specific services. Due to data protection and confidentiality constraints, we do not disclose the identity of the mobile operator or information about the exact location of the data collection campaign. For the same reason, we do not associate performance results to named services. The data collection procedure were conducted under the supervision of the competent national privacy agency, and complies with applicable regulations. In addition, the dataset employed for our study is fully anonymized, as it only provides service traffic information at the antenna without any indication of individual subscriber flows. 

\subsection{Service Usage Overview}

The services that we consider in this study were selected by considering those that \emph{(i)} generate reasonably large amounts of network traffic, \emph{(ii)} represent a variety of application types, and \emph{(iii)} may be a target for network slicing. These considerations ensure that popular high-traffic-volume services for which network resource management is critical are captured. Specifically, our set $\mathcal{S}$ includes 36 distinct services (including YouTube, Netflix, Snapchat, Instagram, Facebook, PokemonGo, Spotify, etc.) that globally are responsible for more than 80\% of the total mobile data traffic in the target region. The services in $\mathcal{S}$ are fairly heterogeneous, and can be categorized into streaming, social media, web, chat, cloud, gaming and miscellaneous classes of applications. This ensures that our selection encompasses services with diverse spatiotemporal dynamics~\cite{shafiq2012,marquez17}, and that our $\mathcal{S}$ provides a relevant case study to evaluate multi-service forecasting in a general, realistic configuration.

An overview of the fraction of the total traffic consumed by each service and each category throughout the duration of the measurement campaign is given in Fig.\,\ref{fig:traffic-stats}. The left plot confirms the power law previously observed in the demands generated by individual mobile services. Also, streaming appears to be the dominant type of traffic, with five services ranking among the top ten. This is confirmed by the right plot, where we can see that streaming accounts for around half of the total traffic consumption. Web, cloud, social media, and chat services also consume large fractions of the total mobile traffic, between 8.47\% and 17.13\%, whereas gaming only accounts for 0.36\% of the demand.

\subsection{Data Preprocessing}
Neural networks that contain convolution operations, including ConvLSTM, naturally require data in matrix form as input \ie regular grid-like structure. However, antennas are non-uniformly distributed in the region covered, which violates the input format requirement. We mitigate this problem by constructing a regular grid that has the same number of points as the number of antennas and subsequently performing a one-to-one antenna-to-point mapping. We do this by minimizing the total geographic distance between antennas and virtual grid points using the Hungarian algorithm~\cite{kuhn1955hungarian}. This translates the antenna point cloud into a gird structure that complies with the matrix form input requirement, while minimizing the spatial displacement and preserving the spatial correlations. Before feeding the measurement data into the neural network, we normalize the traffic volume by subtracting the mean and dividing by the standard deviation. This accelerates the training convergence and leads to superior forecasting performance.

\section{Experiments}
\begin{figure*}[t!]
\centering\includegraphics[width=2\columnwidth]{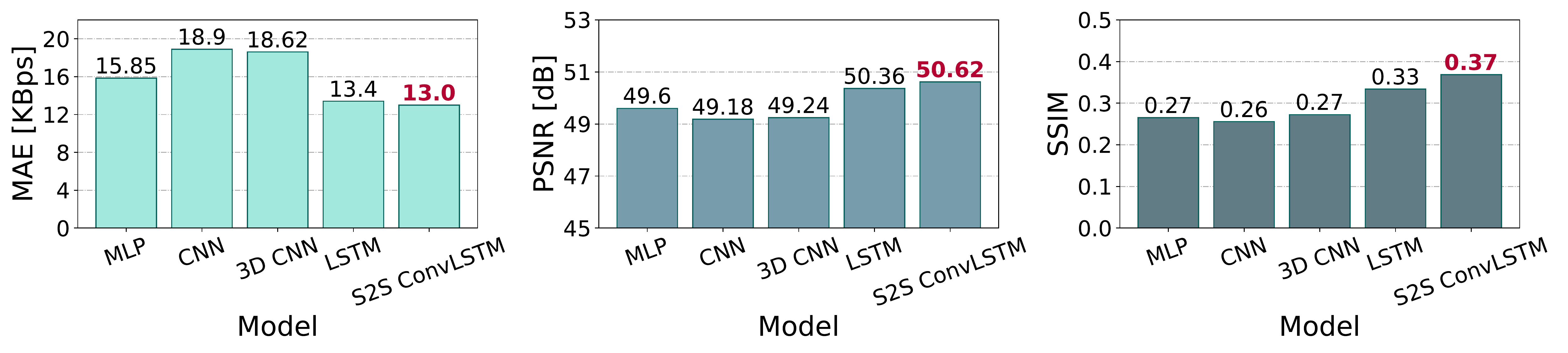}
\caption{Evaluation of all models considered in this study in terms of MAE (left), PSNR (middle) and SSIM (right). The metrics are averaged over the entire test set and the best model is highlighted in bold red.  \label{fig:eva}} 
\end{figure*}

\begin{figure}[t!]
\centering\includegraphics[width=0.95\columnwidth]{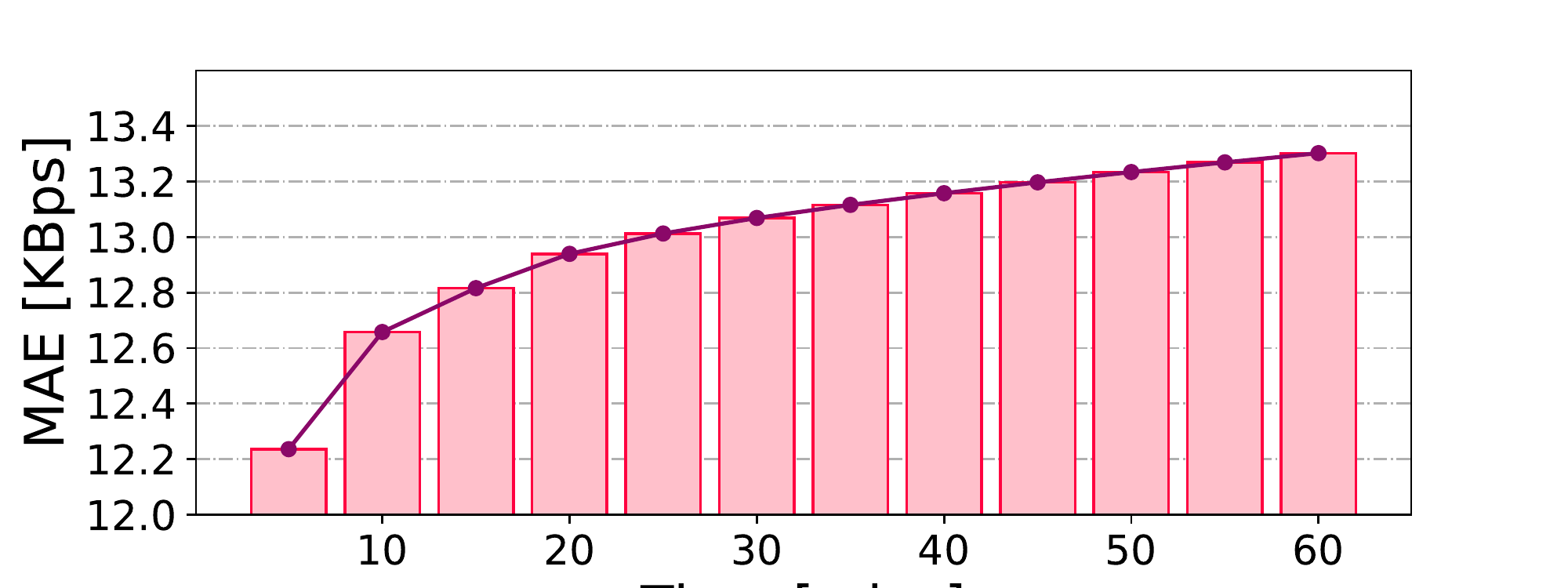}
\caption{The performance of the proposed S2S-ConvLSTM in terms of MAE as the forecasting duration increases. \label{fig:maet}} 
\end{figure}
We build our neural network using the open-source Python libraries TensorFlow~\cite{tensorflow2015-whitepaper} and TensorLayer~\cite{tensorlayer}. The architecture is trained and tested using a high-performance computing cluster with two NVIDIA Tesla K40M GPUs with 2280 cores. In what follows, we first describe discuss benchmarks used for comparison, and the experimental settings and metrics employed for evaluation. Then we report the performance achieved overall and at individual service-level.

\subsection{Benchmarks}

\begin{table}[tb]
\centering
\caption{Benchmark neural networks and their configuration.}
\label{tab:configure}
\begin{tabular}{|C{1.5cm}|C{6.3cm}|}
\hline
\textbf{Model}  & \textbf{Configuration}                                                                                                                            \\ \hline
MLP          & 5 hidden layers, 500 hidden units each.                                                                                       \\ \hline
CNN          & Eleven vanilla convolutional layers with batch normalization layers \cite{ioffe2015batch}, followed by a fully-connected layer.  Each layer is configured with $3\times3$ filters and 128 channels.                                                                  \\ \hline
3D CNN       & Eleven 3D convolutional layers with batch normalization layers \cite{ioffe2015batch}, followed by a fully-connected layer. Each layer is configured with $3\times3\times3$ filters and 128 channels. \\ \hline
LSTM    & Same as Figure \ref{fig:convlstm}, replacing the ConvLSTM blocks with vanilla LSTM blocks with 500 hidden units, and the convolutional layers are replaced by fully-connected layers.                  \\ \hline
\end{tabular}
\end{table}
We compare our proposed ConvLSTM against four different neural network architectures, whose configuration we summarize in Table~\ref{tab:configure}. The multi-layer perceptron (MLP) is the simplest neural network class~\cite{Goodfellow-et-al-2016} with multiple stacks of fully-connected layers. Convolutional neural networks (CNNs) are frequently used for modelling 2D data ~\cite{krizhevsky2012imagenet}. 3D CNN upgrades the CNN by extending the convolution to a third (temporal) dimension. This makes it a good candidate for spatio-temporal data modelling \cite{ji20133d}. Long Short-Term Memory (LSTM) is frequently exploited for modelling sequential data \cite{hochreiter1997long}, such as time series forecasting and natural language processing.

We train all models considered by minimizing the mean square error (MSE) loss function between ground truth observation and predictions, \ie
\begin{equation}\label{eq:mse}
    \mathrm{MSE}(t) = \frac{1}{|\mathcal{S}|\cdot |\mathcal{A}|} \sum_{s \in \mathcal{S}} \sum_{a \in \mathcal{A}}||\widetilde{d}_a^s(t) - d_a^s(t)||^2,
\end{equation}
where $\widetilde{d}_a^s(t)$ denotes the predicted traffic volume at antenna $a$ for service $s$ at time $t$, and $d_a^s(t)$ is the corresponding ground truth. Neural networks are trained using a Stochastic gradient descent (SGD) based optimizer named Adam \cite{kingma2015adam} with the default configuration ($\beta_1 = 0.9, \beta_2 = 0.999, \epsilon = 10^{-8}$) and initial learning rate set to 0.0001. 

\begin{figure*}[htb]
\centering\includegraphics[width=2\columnwidth]{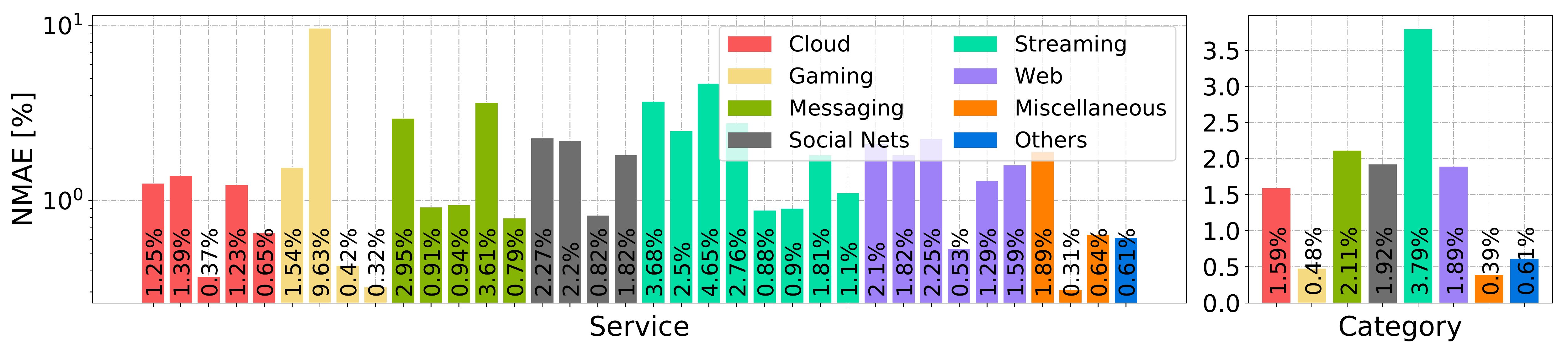}
\vspace*{-1em}
\caption{Prediction performance achieved with the proposed forecasting method in terms of NMAE for all services considered (left) and for 8 service categories that gather multiple services (right). \label{fig:nmae}} 
\end{figure*}

\begin{figure*}[htb]
\centering\includegraphics[width=2\columnwidth]{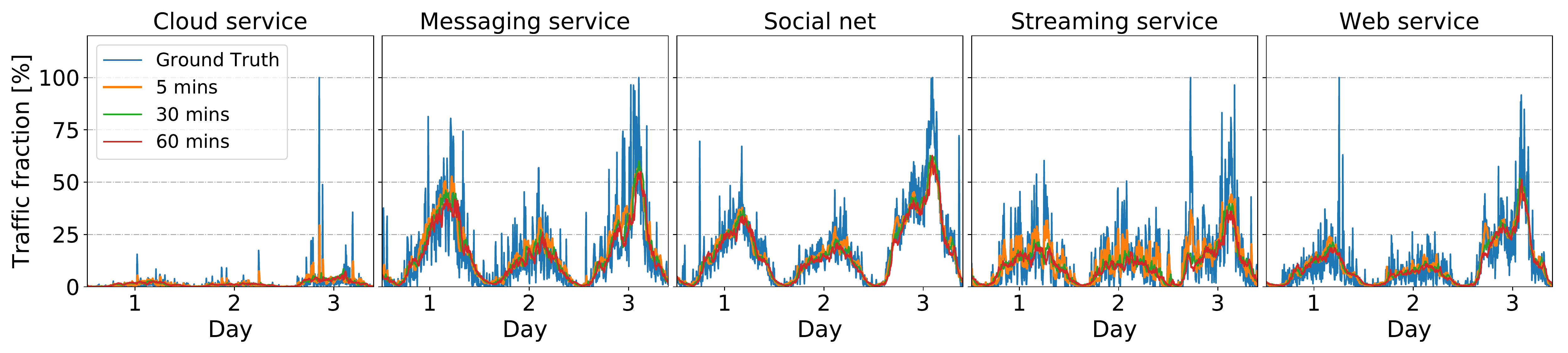}
\vspace*{-1em}
\caption{Example of forecasting results produced by the proposed ConvLSTM on different services, considering  prediction lengths ranging between 5--60 mins (represented by different colors) spanning 3 days, and their associated ground truth. All traffic is normalized to the  traffic peak of the service.   \label{fig:exa}} 
\end{figure*}
\subsection{Experimental Setting and Performance Metrics}
\label{sub:perf}
In our experiments, we employ all neural network models to deliver up to 1-hour long mobile traffic forecasting, given consecutive 1-hour measurement observations taken every 5 minutes. Therefore, all models take as input 12 mobile traffic snapshots and forecast the volume of the following 12 traffic snapshots. Snapshots of different services are regarded as different channels and thus they share the same predictive model. Given that measurements are accumulated every 5 minutes, this means $T = 11$ and $K = 12$. We evaluate the performance of our proposed S2S-ConvLSTM, along with that of all benchmarks, by means of the following three metrics.

The mean absolute error (MAE) is usually used to measure the difference between two variables, but it is also employed as a measure of prediction accuracy:
\begin{equation}
\begin{aligned}
\mathrm{MAE}(t) = \frac{1}{|\mathcal{S}|\cdot |\mathcal{A}|} \sum_{s \in \mathcal{S}} \sum_{a \in \mathcal{A}}|\widetilde{d}_a^s(t) - d_a^s(t)|.
\end{aligned}
\end{equation}

 The peak signal-to-noise ratio (PSNR) is often used to assess the quality of image reconstruction. Nonetheless, this metric has also been employed in networking scenarios, where mobile network traffic snapshots are treated similar to images~\cite{zhang2017zipnet}.
 Formally:
\begin{align}
 \text{PSNR}(t) = &\ 20 \log d_{\max}(t)\ - \nonumber \\
 &\ 10 \log \frac{1}{|\mathcal{S}|\cdot |\mathcal{A}|} \sum_{s \in \mathcal{S}} \sum_{a \in \mathcal{A}}||\widetilde{d}_a^s(t) - d_a^s(t)||^2,
\end{align}
where $d_{\max}(t)$ is the highest traffic volume recorded over all antennas $a\in\mathcal{A}$, and all services $s\in\mathcal{S}$ in our test set.

Structural similarity (SSIM) is traditionally employed for measuring the perceived similarity between uncompressed and compressed images or videos~\cite{hore2010image}. We can take a similar approach to measure the similarity between ground truth traffic snapshots and their predicted counterparts, computing:
\begin{equation}
\begin{aligned}
&\text{SSIM}(t) =  \\
&\frac{\left(2\ \widetilde{d}_a^s(t)\ \mu_d(t) + c_1\right)\left(2\ \text{\sc cov} (d_a^s(t),\widetilde{d}_a^s(t)) + c_2\right)}
{\left(\widetilde{d}_a^s(t)^2\ \mu_d(t)^2  + c_1\right)\left( \text{\sc var} (d_a^s(t)) \text{\sc var}(\widetilde{d}_a^s(t)) +c_2 \right)},
\end{aligned}
\end{equation}
where $\mu_d(t)$ is the average traffic recorded for all services, at all antennas and time instants of the test set. $\text{\sc var}(\cdot)$ and $\text{\sc cov}(\cdot)$ denote the variance and covariance, respectively. Coefficients $c_1$ and $c_2$ are employed to stabilize the fraction in the presence of weak denominators. Following standard practice, we set $c_1 = (k_1L)^2$ and $c_2 = (k_2L)^2$, where $L = 2$ is the dynamic range of float type data, and $k_1 = 0.1$, $k_2 = 0.3$.

\subsection{Performance Comparison}
We summarize the comparative performance evaluation of all the models considered, in terms of MAE, PSNR and SSIM, in Fig.~\ref{fig:eva}. Note that lower MAE, higher PSNR, and higher SSIM indicate better forecasting performance. Observe that the proposed ConvLSTM obtains the best performance across all metrics, which confirms the effectiveness of our proposal. Taking a closer look at Fig.~\ref{fig:eva}, CNN achieves the poorest performance among all the neural network structures studied, due to its lack of modules capable of handling sequential data. 3D CNN mitigates this by extending the convolution operation to the temporal dimension, leading to superior performance over CNN. LSTM works reasonably well in terms of forecasting, as its performance is close to that of the proposed ConvLSTM based solution in terms of all metrics. These results suggest that \emph{(i)} capturing temporal correlation is particularly important for forecasting; indeed LSTM and ConvLSTM work best among all approaches; and \emph{(ii)} the performance of LSTM can be further improved by incorporating convolution operations, as spatial correlations in the geographic distribution of mobile traffic are non-negligible. Overall, our proposal obtains up to 31.2\% higher prediction accuracy, 2.93\% higher fidelity and 37.0\% better structural similarity.

To better appreciate the performance achieved by the proposed S2S-ConvLSTM architecture, in Fig.~\ref{fig:maet} we show the MAE obtained across different prediction steps, \ie when the forecasting length increases. Intuitively, the prediction error will grow with the prediction step, as uncertainty increases with time. Fig.~\ref{fig:maet} confirms our hypothesis, as the MAE grows with the prediction step. Overall, forecasting at 60 minutes ahead yields 8.7\% higher prediction errors than those obtained in the first prediction step (5 minutes ahead), but the errors remain within a reasonable range.

\subsection{Service-wise Evaluation}
Finally, we dive deeper into the performance of the proposed S2S-ConvLSTM, by evaluating the forecasting accuracy for each individual service. To this end, we compute the following normalized MAE (NMAE) associated with a service $s\in\mathcal{S}$:
\begin{equation}
\begin{aligned}
&\mathrm{NMAE}^s = \frac{1}{|\mathcal{A}|\cdot|\mathcal{T}|}\cdot \\ 
&\sum_{a \in \mathcal{A}} \sum_{t \in \mathcal{T}}\frac{|\widetilde{d}_a^s(t) - d_a^s(t)|}{\max_{k \in (t,t+K)} {d_a^s(k)} - \min_{k \in (t,t+K)}{d_a^s(k)}},
\end{aligned}
\end{equation}
where $\mathcal{T}$ denotes the total number of time instances where forecasting is activated. The denominator in the sum terms represents the total span of demand values generated by service $s$ at antenna $a$ from $t+1$ to $t+K$. This normalization makes the performance across services comparable. 

The left plot in Fig.\,\ref{fig:nmae} shows the forecasting performance for each of the 36 services in our reference set $\mathcal{S}$. Observe that 
our S2S-ConvLSTM achieves very accurate performance, as the NMAE values attained are all below 10\%. This suggests that our proposal works well for most of the services. The right plot in Fig.\,\ref{fig:nmae} shows the NMAE performance across different categories that group the same types of services. Observe that steaming services are subject to the highest prediction error, as they consumes a large amount of traffic and exhibit fluctuations frequently. Grouping the same types of services also leads to less noisy traffic streams, and consequently to better forecasting performance. More specifically, the NMAEs evaluated at the category level never exceeds 3.79\% for all categories considered.

For a more detailed view of the performance the proposed S2S-ConvLSTM, in Fig.~\ref{fig:exa} we give the time evolution of the traffic predicted at three times different future steps (\ie 5, 30, and 60 mins ahead, represented by different colors) at the level of a single antenna, for 5 different types of services. Also shown in the plots are the corresponding ground truth observations. Since the time evolution of traffic consumption at antenna level varies significantly, perfectly predicting the exact future consumption becomes impossible. However, the traffic series predictions obtained with our approach overlap remarkably well with the ground truth, irrespective to the forecast length and type of service targeted. This further confirms the effectiveness and generalization ability of our S2S-ConvLSTM approach.

\section{Conclusions}
In this paper, we attacked the multi-service mobile traffic forecasting problem to support resource management for network slicing. We proposed to employ a sequence-to-sequence learning paradigm together with ConvLSTM structures, which can achieve up to 1-hour long traffic forecasting for 36 mobile services with high accuracy, given 1-hour measurement based observations. Experiments conducted over a large-scale mobile traffic dataset collected in an European metropolis demonstrated that our proposal achieves mean absolute error (MAE) performance of 13KBps per antenna, outperforming deep learning benchmarks by up to 31.2\%. Importantly, our solution generalizes well to almost all mobile services since, with the exception of a few outliers, it yields prediction errors below 9.7\% at service level, and only 3.79\% when services are categorized.

\section*{Acknowledgements}
The authors acknowledge the support of the DAUIN High Perforance Computing initiative of Politecnico di Torino, which enabled the training and testing of the neural network architecture designed. Paul Patras also acknowledges the support received from ITU-T through the Focus Group on Machine Learning for Future Networks including 5G. 
{
\bibliographystyle{IEEEtran}
\bibliography{sigproc}

\begin{thebibliography}{10}
\providecommand{\url}[1]{#1}
\csname url@samestyle\endcsname
\providecommand{\newblock}{\relax}
\providecommand{\bibinfo}[2]{#2}
\providecommand{\BIBentrySTDinterwordspacing}{\spaceskip=0pt\relax}
\providecommand{\BIBentryALTinterwordstretchfactor}{4}
\providecommand{\BIBentryALTinterwordspacing}{\spaceskip=\fontdimen2\font plus
\BIBentryALTinterwordstretchfactor\fontdimen3\font minus
  \fontdimen4\font\relax}
\providecommand{\BIBforeignlanguage}[2]{{%
\expandafter\ifx\csname l@#1\endcsname\relax
\typeout{** WARNING: IEEEtran.bst: No hyphenation pattern has been}%
\typeout{** loaded for the language `#1'. Using the pattern for}%
\typeout{** the default language instead.}%
\else
\language=\csname l@#1\endcsname
\fi
#2}}
\providecommand{\BIBdecl}{\relax}
\BIBdecl

\bibitem{cisco2017}
{Cisco}, ``{Cisco Visual Networking Index: Forecast and Methodology,
  2017-2022},'' February 2019.

\bibitem{bega2019deepcog}
D.~Bega, M.~Gramaglia, M.~Fiore, A.~Banchs, and X.~Costa-Perez, ``{DeepCog}:
  Cognitive network management in sliced 5g networks with deep learning,'' in
  \emph{INFOCOM 2019}.

\bibitem{naboulsi2016large}
D.~Naboulsi, M.~Fiore, S.~Ribot, and R.~Stanica, ``Large-scale mobile traffic
  analysis: a survey,'' \emph{IEEE Communications Surveys \& Tutorials},
  vol.~18, no.~1, pp. 124--161, 2016.

\bibitem{keysight}
{Keysight Technologies}, ``{J7523A Cellular Network Passive Probe -- Data
  Sheet},'' accessed April 2017, 2017.

\bibitem{zhang2018deep}
C.~Zhang, P.~Patras, and H.~Haddadi, ``Deep learning in mobile and wireless
  networking: A survey,'' \emph{IEEE Communications Surveys \& Tutorials},
  2019.

\bibitem{zhang2017zipnet}
C.~Zhang, X.~Ouyang, and P.~Patras, ``{ZipNet}-{GAN}: Inferring fine-grained
  mobile traffic patterns via a generative adversarial neural network,'' in
  \emph{Proc. ACM CoNEXT}, 2017, pp. 363--375.

\bibitem{zhang2018long}
C.~Zhang and P.~Patras, ``Long-term mobile traffic forecasting using deep
  spatio-temporal neural networks,'' in \emph{Proc. ACM MobiHoc}, 2018.

\bibitem{wang2018spatio}
X.~Wang, Z.~Zhou, F.~Xiao, K.~Xing, Z.~Yang, Y.~Liu, and C.~Peng,
  ``Spatio-temporal analysis and prediction of cellular traffic in
  metropolis,'' \emph{IEEE Transactions on Mobile Computing}, 2018.

\bibitem{sutskever2014sequence}
I.~Sutskever, O.~Vinyals, and Q.~V. Le, ``Sequence to sequence learning with
  neural networks,'' in \emph{Advances in neural information processing
  systems}, 2014, pp. 3104--3112.

\bibitem{xingjian2015convolutional}
S.~Xingjian, Z.~Chen, H.~Wang, D.-Y. Yeung, W.-K. Wong, and W.-c. Woo,
  ``Convolutional lstm network: A machine learning approach for precipitation
  nowcasting,'' in \emph{Advances in neural information processing systems},
  2015, pp. 802--810.

\bibitem{shafiq2012}
M.~Z. Shafiq, L.~Ji, A.~X. Liu, J.~Pang, and J.~Wang, ``Characterizing
  geospatial dynamics of application usage in a {3G} cellular data network,''
  in \emph{Proc. IEEE INFOCOM}, March 2012, pp. 1341--1349.

\bibitem{marquez17}
C.~Marquez, M.~Gramaglia, M.~Fiore, A.~Banchs, C.~Ziemlicki, and Z.~Smoreda,
  ``Not all apps are created equal: Analysis of spatiotemporal heterogeneity in
  nationwide mobile service usage,'' in \emph{Proc. ACM CoNEXT}, 2017.

\bibitem{kuhn1955hungarian}
H.~W. Kuhn, ``The {Hungarian} method for the assignment problem,'' \emph{Naval
  Research Logistics (NRL)}, vol.~2, no. 1-2, pp. 83--97, 1955.

\bibitem{tensorflow2015-whitepaper}
M.~Abadi, P.~Barham, J.~Chen, Z.~Chen, A.~Davis, J.~Dean, M.~Devin,
  S.~Ghemawat, G.~Irving, M.~Isard \emph{et~al.}, ``{TensorFlow}: A system for
  large-scale machine learning.'' in \emph{OSDI}, vol.~16, 2016, pp. 265--283.

\bibitem{tensorlayer}
H.~Dong, A.~Supratak, L.~Mai, F.~Liu, A.~Oehmichen, S.~Yu, and Y.~Guo,
  ``{TensorLayer}: A versatile library for efficient deep learning
  development,'' in \emph{Proc. ACM Multimedia}, 2017, pp. 1201--1204.

\bibitem{ioffe2015batch}
S.~Ioffe and C.~Szegedy, ``Batch normalization: Accelerating deep network
  training by reducing internal covariate shift,'' in \emph{International
  Conference on Machine Learning}, 2015, pp. 448--456.

\bibitem{Goodfellow-et-al-2016}
I.~Goodfellow, Y.~Bengio, and A.~Courville, \emph{Deep Learning}.\hskip 1em
  plus 0.5em minus 0.4em\relax MIT Press, 2016,
  \url{http://www.deeplearningbook.org}.

\bibitem{krizhevsky2012imagenet}
A.~Krizhevsky, I.~Sutskever, and G.~E. Hinton, ``Imagenet classification with
  deep convolutional neural networks,'' in \emph{NIPS}, 2012, pp. 1097--1105.

\bibitem{ji20133d}
S.~Ji, W.~Xu, M.~Yang, and K.~Yu, ``3{D} convolutional neural networks for
  human action recognition,'' \emph{IEEE transactions on pattern analysis and
  machine intelligence}, vol.~35, no.~1, pp. 221--231, 2013.

\bibitem{hochreiter1997long}
S.~Hochreiter and J.~Schmidhuber, ``Long short-term memory,'' \emph{Neural
  computation}, vol.~9, no.~8, pp. 1735--1780, 1997.

\bibitem{kingma2015adam}
D.~Kingma and J.~Ba, ``{Adam: A method for stochastic optimization},''
  \emph{International Conference on Learning Representations (ICLR)}, 2015.

\bibitem{hore2010image}
A.~Hore and D.~Ziou, ``Image quality metrics: {PSNR vs. SSIM},'' in \emph{Proc.
  IEEE International Conference on Pattern Recognition (ICPR)}, 2010, pp.
  2366--2369.

\end{thebibliography}
}

\end{document}